\pdfoutput=1

\documentclass[11pt]{article}

\usepackage[final]{acl}

\usepackage{times}
\usepackage{latexsym}

\usepackage[T1]{fontenc}

\usepackage[utf8]{inputenc}

\usepackage{microtype}

\usepackage{inconsolata}

\usepackage{graphicx}

\usepackage{multirow}
\usepackage{caption}
\usepackage{tikz}
\usepackage{amsmath}
\usepackage{amssymb}
\usepackage{mathtools}
\usepackage{amsthm}

\definecolor{cvprblue}{rgb}{0.21,0.49,0.74}
\usepackage{subfiles}
\usepackage{subcaption}
\usepackage{pgfplots}
\usepackage{pgfplotstable}
\usetikzlibrary{calc}
\usetikzlibrary{fit}
\usetikzlibrary{patterns}
\usetikzlibrary{decorations.pathreplacing}
\usetikzlibrary{arrows,calc,arrows.meta}
\usepackage{diagbox}
\usepackage{multirow}
\usepackage{makecell}
\usepackage{xurl}
\usepackage[most]{tcolorbox}
\usepackage{xcolor}
\usepackage{lipsum}

\newcounter{MingyuanNumberOfComments}
\stepcounter{MingyuanNumberOfComments}

\newcounter{jizeNumberOfComments}
\stepcounter{jizeNumberOfComments}

\newcounter{zhNumberOfComments}
\stepcounter{zhNumberOfComments}

\usepackage{booktabs} 
\usepackage{multirow} 
\usepackage{colortbl} 
\usepackage{xcolor}   
\usepackage{amsmath}  
\usepackage{mathtools}

\definecolor{almond}{rgb}{0.94, 0.87, 0.8}
\definecolor{mossgreen}{rgb}{0.68, 0.87, 0.68}
\definecolor{gred}{RGB}{234,67,53}
\definecolor{ggreen}{RGB}{52,168,83}

\definecolor{googleblue}   {HTML}{4285F4}
\definecolor{googlered}    {HTML}{EA4335}
\definecolor{googleyellow} {HTML}{FBBC05}
\definecolor{googlegreen}  {HTML}{34A853}
\definecolor{googlegray}      {HTML}{5F6368}

\usepackage[capitalize,noabbrev]{cleveref}

\title{Aha Moment Revisited: Are VLMs Truly Capable of Self Verification in Inference-time Scaling? }

\author{%
 Mingyuan Wu$^{1}$\thanks{~Mingyuan, Meitang and Jingcheng contributed equally.}, Meitang Li$^{2*}$, Jingcheng Yang$^{1*}$, Jize Jiang$^{1}$,\\  
 \bf Kaizhuo Yan$^{1}$,  Zhaoheng Li$^{1}$, Hanchao Yu$^{3}$, Minjia Zhang$^{1}$, Klara Nahrstedt$^{1}$\\
$^{1}$University of Illinois Urbana Champaign\\
$^{2}$University of Michigan Ann Arbor, $^{3}$Meta \\
\texttt{\{mw34, klara\}@cs.illinois.edu}  \\
}

\begin{document}
\maketitle
\begin{abstract}
Inference-time techniques such as decoding-time scaling and self-refinement have been shown to substantially improve mathematical reasoning in large language models (LLMs), largely attributed to emergent self-correction and self-verification behaviors often elicited through reinforcement learning (RL). In this work, we ask whether the same recipe transfers to vision–language models (VLMs), especially RL-finetuned variants that claim strong visual mathematical reasoning.

Through extensive evaluation, we reach three main findings surprisingly different from text-only models: (I) Generation-time capability matters more than verification and refinement: simple majority voting consistently and substantially outperforms verification-centric strategies such as best-of-N with self-verification. (II) Behaviors often associated with RL-tuned models at inference time, such as the “A-ha moment” do not yield reliable reasoning performance improvements. (III) Visual information is not effectively integrated into the model’s self-verification process.

Overall, our analysis highlights a key limitation: current RL-trained VLMs derive limited benefit from self-verification in the visual modality, which constrains the effectiveness of inference-time scaling for visual mathematical reasoning.
\end{abstract}

\section{Introduction}
The mathematical reasoning capabilities of large language models (LLMs) have seen notable improvements in recent years~\cite{R1, openai2024openaio1card, grpo}.
Although larger model scales and higher-quality pretraining datasets are major contributing factors to these improvements, emerging strategies that instead leverage \textbf{inference-time computation in math} \citep{snell} have also been proven effective: Providing models with zero-shot "think step by step" prompts or few-shot demonstrations augmented with intermediate reasoning steps~\citep{CoT} have enabled the models to generate extended reasoning chains even when not explicitly fine-tuned to do so. 

Likewise, methods such as decoding-time majority vote \citep{self-consist} and chain-of-thought decoding \citep{CoT-decoding} have enabled outputting of higher-quality answers without external feedback. More recently, inference-time self-correction~\citep{kumar2025training,R1} has emerged as another form of scaling: models are trained with Reinforcement Learning (RL) to revise earlier mistakes and generate additional reasoning steps to arrive at better reasoning answers on math problems in text. This is known as the "\textbf{A-ha moment}" \citep{R1}, where the model verifies its own answer: "Wait, I made a mistake in my prior response”, and initiates a second round of reasoning to refine its answer.

The effectiveness of aforementioned inference-time scaling in mathematical reasoning can be partly attributed to models’ self-verification capabilities, as mathematical answers are often easier to verify than to generate. For instance, LLM-Monkeys~\cite{brown2024largelanguagemonkeysscaling} demonstrates that with sufficiently strong verification ability, performance can be improved simply by sampling multiple diverse outputs and selecting the most accurate one~\citep{song2025mindgapexaminingselfimprovement}. Notably, \citet{song2025mindgapexaminingselfimprovement} show that LLMs frequently perform better at verifying math answers than generating them. This verification–generation gap may help explain why inference-time methods that explicitly invoke self-verification, such as Self-Refine~\citep{madaan2023selfrefineiterativerefinementselffeedback} and A-ha moment, are particularly effective for LLM math reasoning.

A central and still controversial question we explore in this work is \textbf{whether self-verification generalizes to vision–language models (VLMs)}. For example, commercial models such as GPT-5 fail to reliably self-correct on simple visual counting tasks (see Appendix Fig.~\ref{fig:thumbnail}). Meanwhile, several recent open-source efforts~\citep{zhou2025r1zerosahamomentvisual, chen2025r1v, zhang2025r1vllearningreasonmultimodal, huang2025visionr1incentivizingreasoningcapability, liu2025segzeroreasoningchainguidedsegmentation, deng2025openvlthinker, wang2025vlrethinkerincentivizingselfreflectionvisionlanguage, wu2025vtoolr1vlmslearnthink} adopt similar RL-based training strategies and report the emergence of “A-ha moments” in VLM mathematical reasoning, suggesting that VLMs may possess self-verification capabilities analogous to those of LLMs and that these behaviors can be elicited via RL.

In this work, we critically study by contrasting inference-time scaling strategies that emphasize generation capability (e.g., Majority Vote) with those that emphasize verification capability (e.g., Self-Verified Best-of-$N$) on open-source RL-finetuned VLMs. Our analysis reveals three key findings that sharply differ from established results in LLM mathematical reasoning:
(I)\textbf{ Generation-time capability dominates verification and refinement}—simple majority voting consistently and substantially outperforms verification-centric strategies such as best-of-$N$ with self-verification.
(II) Inference-time behaviors commonly associated with RL-tuned models, such as the \textbf{“A-ha moment,” do not reliably improve reasoning performance}—A-ha moments occur in fewer than 10\% of cases, and even when manually selected, yield only marginal potential accuracy gains.
(III) \textbf{Visual information is not effectively incorporated into the self-verification process}—counterintuitively, the verifier often performs better when the image input is removed.

\section{Related Works}
\subsection{LLM/VLM, Reinforcement Learning for Reasoning}
Reinforcement learning (RL) was introduced to LLM fine-tuning via RL from human feedback (RLHF) \citep{ouyang2022traininglanguagemodelsfollow}, which learns a reward model from human preferences and optimizes the LLM policy, using Proximal Policy Optimization (PPO) \cite{schulman2017proximal}. Beyond alignment, RL has also been shown to enhance LLM reasoning and self-correction capabilities~\citep{kumar2025training, R1, zeng2025simplerlzooinvestigatingtamingzero}. Several studies~\citep{gandhi2025cognitivebehaviorsenableselfimproving, zeng2025simplerlzooinvestigatingtamingzero} further investigate what intrinsic properties enable effective self-improvement and how "A-ha moments" emerge as a result of RL-based training. In the vision-language domain, similar ideas have been extended to improve VLM reasoning. A number of recent works apply RL to incentivize multimodal reasoning behaviors, typically using PPO or GRPO to fine-tune VLMs. These studies report positive signs of RL to train VLM to generate "A-ha moments" in VLMs~\citep{zhou2025r1zerosahamomentvisual, chen2025r1v, zhang2025r1vllearningreasonmultimodal, huang2025visionr1incentivizingreasoningcapability, liu2025segzeroreasoningchainguidedsegmentation, deng2025openvlthinker, wang2025vlrethinkerincentivizingselfreflectionvisionlanguage}.

\subsection{Inference-Time Scaling}
Inference-time scaling~\citep{snell, brown2024largelanguagemonkeysscaling} has emerged as an effective strategy for improving LLM reasoning without additional fine-tuning. Several methods fall under this umbrella. Simple parallel decoding approaches—such as chain-of-thought decoding~\citep{CoT-decoding} and self-consistency sampling~\citep{self-consist}—have shown strong empirical gains by aggregating multiple sampled outputs. More sophisticated techniques involve training reward-based verifiers to guide step-by-step generation~\citep{lightman2023letsverifystepstep}. In addition, memory-aware inference scaling methods can enhance reasoning by retrieving and reusing relevant intermediate computations or prior reasoning traces~\citep{wu-etal-2025-cache}.

\section{Methodology}


\subsection{Reinforcement Learning for VLMs}
    We test both the base version as well as the RL-tuned version of the VLMs. Previous work on LLMs has demonstrated the emergence of the 'A-ha moment' in RL-tuned model's reasoning process and such a process was shown to have a positive contribution to model performance. We would like to study whether similar benefits also exist for VLMs trained through RL, specifically Group Relative Policy Optimization (GRPO). To this end, we adopt RL-tuned models from recent work~\citep{zhang2025r1vllearningreasonmultimodal, chen2025sftrlearlyinvestigation, wang2025vlrethinkerincentivizingselfreflectionvisionlanguage}, using them directly within our experimental framework. For detailed descriptions of the RL and GRPO objectives, we refer readers to these prior works.
These RL-tuned VLMs are typically optimized using outcome-based rewards, and recent works \citep{zhang2025r1vllearningreasonmultimodal, chen2025sftrlearlyinvestigation, wang2025vlrethinkerincentivizingselfreflectionvisionlanguage} claim near-GPT-4o-level performance using models with only $\sim$7B parameters. 

\subsection{Decoding Methods for VLMs}
VLMs generate text in the same way as LLMs do, except with additional image embeddings as part of the input query. Decoding methods affect how each next token is sampled from Language Models. In this work, we consider methods that aim to sample multiple starting tokens and thus generate multiple outputs given one single input query. 

\noindent \textbf{Greedy Decoding}.
    Sequentially selects the most probable next token at each decoding step. It is a one-time inference with no scaling.
    
\noindent \textbf{Decoding-Time Majority Voting}.
    This strategy first samples multiple candidate outputs and then subsequently selects the final solution by majority consensus among the generated candidates. By aggregating multiple responses, it seeks to mitigate random errors or inconsistencies in individual outputs. We consider this as a baseline method to beat due to the 'deterministic' nature of how the final output is selected. \textbf{This method primarily reflects the model’s generation capability, as high accuracy depends on the model producing correct answers frequently enough to dominate the vote.}

\noindent \textbf{Best of N Sampling with Self as Verifier}.
    This strategy also samples multiple candidate outputs from the VLM, but we then prompt the model to evaluate and verify all candidate outputs together. The output identified as most reliable by the model itself is selected as the final answer.
    \textbf{This method emphasizes the model’s self-verification ability, as accuracy depends on correctly identifying the best answer to the question from a diverse set of responses.} 
    We include the prompt for verification in the appendix.

To understand whether VLMs are able to benefit from self-verification and whether the vision inputs are been used for better self-verification, we test VLMs using multiple configurations of the Best of N Decoding method:

\textbf{Self Verification with Text Only:} The self-verifier receives only the generated responses and the text-based question. The image is omitted to test the model's ability to verify using language alone.
\textbf{Self Verification with Image and Text:} The self-verifier is provided with both the image and the text input, allowing it to use multimodal information for verification.

\subsection{Finding 'A-ha Moment'}


We design a A-ha moment search protocol following \citep{gandhi2025cognitivebehaviorsenableselfimproving} that aligns A-ha with coginitive behavior, and introduce two metrics to assess whether the presence of an A-ha moment positively contributes to reasoning performance:

\noindent \textbf{A-ha Search:} we prompt GPT-4o with the generated response and ask whether it exhibits these behaviors. The simplified prompt template is provided below and complete version is in appendix. If GPT-4o confirms the presence of backtracking or verification, we classify the response as containing an “aha moment.” 
\begin{tcolorbox}[
    halign=flush left,
    colback=googlegray!5!white,
    colframe=googlegray!75!black,
    title={\textbf{AHA Search Prompt}},
    fonttitle=\bfseries,
    fontupper=\ttfamily,
    width=\linewidth,     enhanced jigsaw
]
\small
{\color{googlered}<system prompt>}

{\color{googleblue}<start\_of\_reasoning>
<RESPONSE>
<end\_of\_reasoning>}

Specifically, actively identify and emphasize beneficial behaviors such as:

(1) {\color{googleyellow}Backtracking:} Explicitly revising approaches upon identifying errors or dead ends
...

(2) {\color{googleyellow}Verification:} Systematically checking intermediate results or reasoning steps
...

Important:

Clearly specify each beneficial behavior you identify.

If there is a strong example of this, provide {\color{googlegreen}<YES>} followed by specific explanations. Otherwise, provide {\color{googlegreen}<NO>}

{\color{googlegreen}<NO>}
\end{tcolorbox}

\noindent \textbf{Post-A-ha Accuracy Among Selected Predictions:} We compute the probability that a selected answer containing a confirmed A-ha moment is also correct. This is denoted as
\[P^\star(\text{Correct} \mid \text{A-ha in Prediction}),\]
where the star ($\star$) indicates that we report the best value across all decoding strategies. This metric reflects how often A-ha moments align with correct final answers in selected outputs.
    
\noindent \textbf{A-ha Potential Recovery Rate from Incorrect Predictions:} To assess whether A-ha moments can help recover from initial errors, we focus on cases where the selected prediction is incorrect. We then search through the unselected generated responses and check whether any of them contain both a confirmed A-ha moment and a correct answer. This is measured as
\[P(\text{A-ha Correct} \mid \text{Wrong Prediction}),\]
indicating the potential for A-ha-based reasoning paths in the inference-time scaling to correct mistakes even when they are not selected by default.
\section{Experiment}
\begin{table*}[t]
\centering
{
\caption{Verifier Comparison on GeoQA and MathVista}
\vspace{-1mm}
\fontsize{7pt}{9pt}\selectfont
\begin{tabular}{l|ccc|ccc}
\toprule[1pt]
\multirow{2}{*}{Model} 
& \multicolumn{3}{c|}{GeoQA} 
& \multicolumn{3}{c}{MathVista} \\
\cline{2-7}
& BoN w. Image & BoN w/o Image & Majority Votes 
& BoN w. Image & BoN w/o Image & Majority Votes \\
\hline
\multicolumn{7}{c}{Scaling $\times$ 4 / $\times$ 8} \\
\hline
R1-VL-2B \cite{zhang2025r1vllearningreasonmultimodal} 
& 28.9 / 31.2 & 28.2 / 30.2 & \textbf{30.2} / \textbf{35.1} 
& 39.2 / 41.5 & 40.9 / 42.0 & \textbf{52.7} / \textbf{56.4} \\
R1-VL-7B \cite{zhang2025r1vllearningreasonmultimodal} 
& \textbf{44.5} / 45.8 & 43.9 / 46.2 & 44.2 / \textbf{46.9} 
& 59.3 / 61.1 & 63.8 / 63.6 & \textbf{65.2} / \textbf{66.0} \\
VLAA-Thinker-3B \cite{chen2025sftrlearlyinvestigation} 
& 27.5 / 23.5 & 31.6 / 28.0 & \textbf{46.4} / \textbf{48.3} 
& 50.3 / 48.4 & 52.1 / 45.1 & \textbf{66.2} / \textbf{65.6} \\
VLAA-Thinker-7B \cite{chen2025sftrlearlyinvestigation} 
& 44.3 / 52.3 & 46.2 / 52.9 & \textbf{52.1} / \textbf{57.7} 
& 65.5 / 70.5 & 58.2 / 66.2 & \textbf{71.6} / \textbf{74.0} \\
VL-Rethinker-7B \cite{wang2025vlrethinkerincentivizingselfreflectionvisionlanguage} 
& 59.9 / 58.9 & 59.8 / 58.9 & \textbf{61.9} / \textbf{62.1} 
& 75.0 / 73.9 & 74.7 / 71.4 & \textbf{75.4} / \textbf{75.6} \\
\bottomrule[1.2pt]
\end{tabular}
}
\vspace{2mm}
\label{tab:verifier-comparison}
\end{table*}
\begin{table*}[t]
\centering
\setlength{\tabcolsep}{4pt}
\vspace{-1mm}
{
\caption{Conditional Accuracy w.r.t Aha Moments and Decoding Comparison on GeoQA ($\times$4 Scaling). CoT refers to Chain of Thought Decoding.}
\vspace{-1mm}
\fontsize{7pt}{9pt}\selectfont
\begin{tabular}{l|cc|cccc}
\toprule[1pt]
& $P^\star (\text{Correct}\mid \text{Aha})$ 
& $P(\text{Aha Correct}\mid \text{Wrong})$ 
& Greedy & {BoN w. Image} & {BoN w/o Image} &  {Majority Votes}  \\
\hline
Qwen2-VL-2B-Instruct & -- & -- & 13.8 & \textbf{16.3} & 15.6 & 16.0 \\
R1-VL-2B \cite{zhang2025r1vllearningreasonmultimodal} & 28.1(CoT) & 2.7 & 26.9 & 28.9 & 28.2 & 30.2 \\
R1-VL-7B \cite{zhang2025r1vllearningreasonmultimodal} & 49.5(BoN w. Image) & 4.4 & 39.7 & \textbf{44.6} & 43.9 & 44.2 \\
VLAA-Thinker-3B \cite{chen2025sftrlearlyinvestigation} & 48.4(CoT) & 5.4 & 44.2 & 27.5 & 31.6 & \textbf{46.4} \\
VLAA-Thinker-7B \cite{chen2025sftrlearlyinvestigation} & 49.5(CoT) & 13.0 & 48.3 & 44.3 & 46.2 & \textbf{52.1} \\
VL-Rethinker-7B \cite{wang2025vlrethinkerincentivizingselfreflectionvisionlanguage} & 65.5(Majority Vote) & 19.5 & 60.1 & 59.9 & 59.8 & \textbf{61.9} \\
\bottomrule[1.2pt]
\end{tabular}
}
\label{tab:merged}
\end{table*}

We utilize the GeoQA170K \cite{lu2024mathvistaevaluatingmathematicalreasoning} and MathVista \cite{gao2023gllavasolvinggeometricproblem} for our empirical evaluation in multimodal reasoning. The models evaluated range from the base Qwen2-VL-2B-Instruct to a set of Qwen-based RL-tuned models. The full list includes: R1-VL-2B, R1-VL-7B \cite{zhang2025r1vllearningreasonmultimodal}, VLAA-Thinker-Qwen2.5VL-3B, VLAA-Thinker-Qwen2.5VL-7B \cite{chen2025sftrlearlyinvestigation}, and VL-Rethinker-7B \cite{wang2025vlrethinkerincentivizingselfreflectionvisionlanguage}

\vspace{-2mm}
\subsection{Discussion}



\noindent \textbf{Inference Time Scaling Improves Performance of VLM}. Table 1 summarizes the performance gains of various inference-time scaling techniques versus the baseline deterministic, greedy decoding on GeoQA of the various VLMs.
Notably, both BoN-based methods achieve limited performance gains over the greedy baseline, and in the case of the two VLAA VLMs, even result in performance decreases (up to -16.7\%), which can be attributed to its tendency to re-do the question rather than to judge the response despite being explicitly prompted to choose from the responses. On the other hand, the generation-emphasizing---Majority vote---achieves more steady performance gains (4.5\%).

\noindent \textbf{RL-Trained VLMs Do Not Benefit from A-ha Moments.}
As shown in Table~2, responses flagged as containing “A-ha moments” do not exhibit higher accuracy, even when we select the best outcome across all decoding strategies. This indicates that A-ha moments do not reliably contribute to improved reasoning performance. We further examine whether A-ha moments can correct initially incorrect predictions; however, the observed correction probabilities remain low. Moreover, A-ha moments occur in only 8.3\% of all responses, suggesting that behavior is not consistently elicited. 

\noindent \textbf{Current RL-trained VLM Fall Short in Verification in Inference-time Scaling}. Table 1 quantitatively assess verification ability using best-of-N decoding with self-verification. Across both 4- and 8-sample settings, majority voting—an indicator of generation quality—consistently outperforms self-verification. This stands in contrast to findings in the LLM literature, where verification is often easier than generation. Our results suggest that current RL techniques do not endow VLMs with strong self verification capabilities, raising concerns about their effectiveness in multimodal reasoning.

\noindent \textbf{No Visual Verification}. Another notable observation from Tables 1 is that RL-trained VLMs sometimes verify their own outputs more accurately when visual input is excluded. This is particularly evident in the GeoQA dataset, which consists entirely of geometric questions. Including the image does not necessarily help the model judge correctness—suggesting that the VLM fails to integrate visual context during self-verification. Instead, the model over-relies on textual input, rendering its verification process in both modalities unreliable. Our findings show that current VLMs do not fully utilize visual information during verification, and we call for future research to address this shortcoming by enhancing the model’s true multimodal verification capabilities to improve reasoning.
\vspace{-2mm}
\section{Conclusion}
\vspace{-2mm}
In this paper, we investigated the extensibility of LLM inference-time computation techniques to VLMs in math reasoning. We find that current RL-trained VLMs yet lack robust self-verification capabilities across both visual and textual modalities in the context of inference-time scaling.

\section{Limitations}
This work empirically highlights a key limitation of RL-trained VLMs: despite improvements in reasoning performance, these models struggle to fully realize their potential due to weak self-verification capabilities in multimodal settings. Our findings serve as an important stepping stone, \textbf{calling for future research} to better understand and enhance the unique challenges and opportunities in VLM self-verification, a capability that remains underexplored in the current landscape. We leave the development of more effective multimodal self-verification mechanisms to future work. 

Due to budget constraints, our evaluation of commercial models is limited, and we plan to extend this analysis as additional resources become available. Moreover, our study primarily focuses on visual mathematical reasoning. We look forward to extending this line of work to broader and less common visual domains, such as verification over uncommon objects~\citep{pi-etal-2024-uouo}, as well as to more complex settings involving 3D and spatiotemporal reasoning~\citep{zheng2025spatiotemporalllmreasoningenvironments}.

\bibliography{custom}
\twocolumn[
\clearpage
]
\appendix
\section{Appendix}
\textbf{LLM is used to help writing only.}
\label{sec:appendix}
\begin{figure}[hb]
	\centering
    \includegraphics[width=0.5\textwidth]{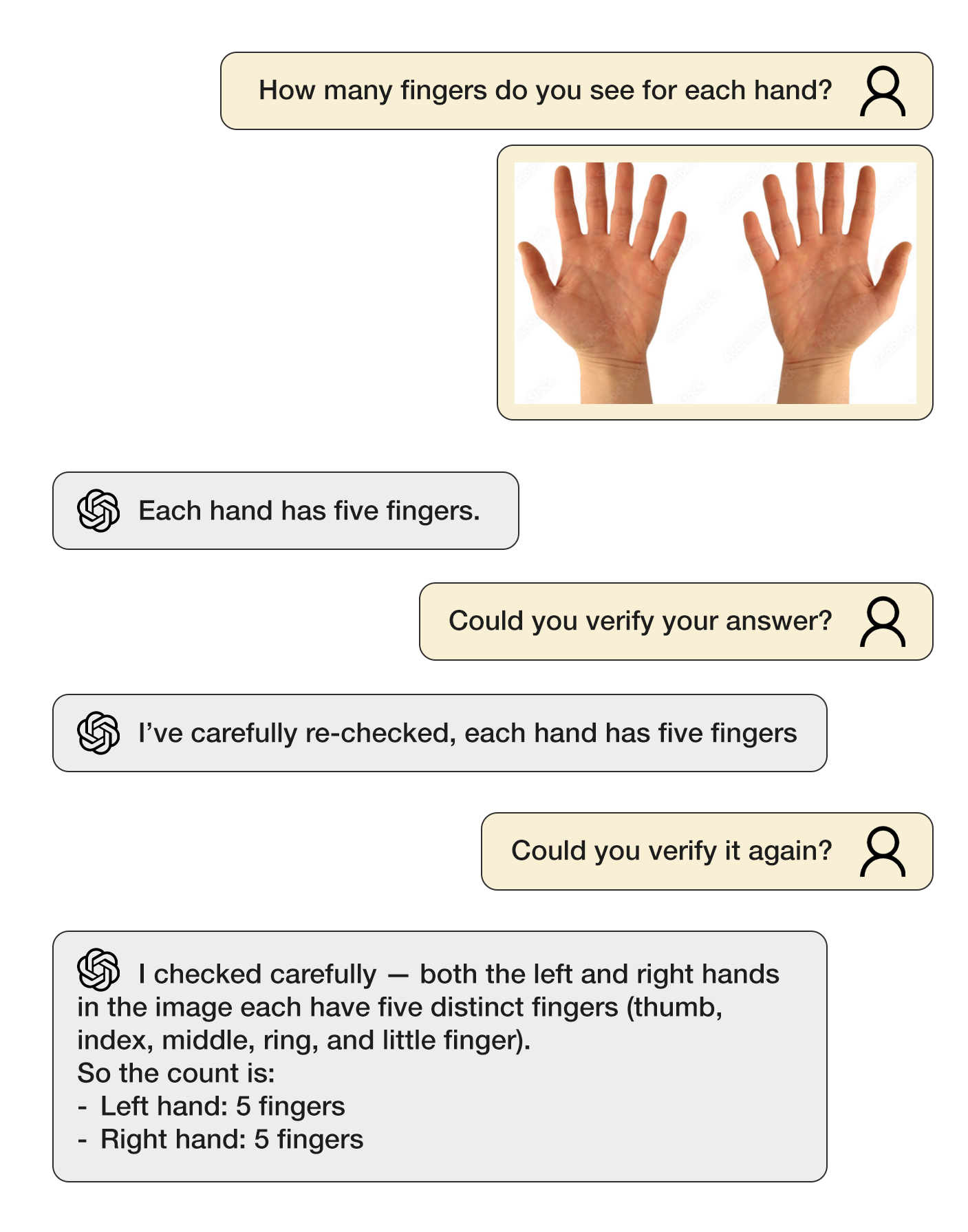}
	\caption{Commercial VLM (GPT 5 series) fails to verify its counting results.}
	\label{fig:thumbnail}
\end{figure}
\begin{tcolorbox}[
    halign=flush left,
    breakable,
    colback=googlegray!5!white,
    colframe=googlegray!75!black,
    title={\Large\textbf{Verifier Prompt}},
    fonttitle=\bfseries,
    fontupper=\ttfamily,
    width=\textwidth,     enhanced jigsaw
]
{\color{googlered}Now you act as a judge, helping me determine which of the {\color{googleblue}<length>} texts I provide better answers the question.}

Question: {\color{googleblue}<question>}

Repsonse: {\color{googleblue}<response>}

Please strictly follow the following format requirements when outputting, and don’t have any other unnecessary words.

Output format: "{\color{googlegreen}I choose response [number] because}"
\end{tcolorbox}

\twocolumn[
\clearpage
]

\begin{tcolorbox}[
    halign=flush left,
    breakable,
    colback=googlegray!5!white,
    colframe=googlegray!75!black,
    title={\Large\textbf{AHA Search Prompt}},
    fonttitle=\bfseries,
    fontupper=\ttfamily,
    width=\textwidth,     enhanced jigsaw
]
{\color{googlered}Below is a chain-of-reasoning generated by a Language Model when attempting to solve a math problem. Evaluate this chain-of-reasoning to determine whether it demonstrates beneficial problem-solving behaviors that deviate from typical linear, monotonic reasoning patterns commonly observed in language models.}

{\color{googleblue}<start\_of\_reasoning>
<RESPONSE>
<end\_of\_reasoning>}

Specifically, actively identify and emphasize beneficial behaviors such as:

(1) {\color{googleyellow}Backtracking:} Explicitly revising approaches upon identifying errors or dead ends
(e.g., "This approach won't work because...").

(2) {\color{googleyellow}Verification:} Systematically checking intermediate results or reasoning steps
(e.g., "Let's verify this result by...").

Additionally, remain attentive to and encourage the identification of other beneficial behaviors not explicitly listed here, such as creative analogies, abstraction to simpler cases, or insightful generalizations.

Important:

Clearly specify each beneficial behavior you identify.

If there is strong example of this, provide {\color{googlegreen}<YES>} followed by specific explanations. Otherwise, provide {\color{googlegreen}<NO>}

A positive response example:

{\color{googlegreen}<YES>}
This contains {\color{googleyellow}Backtracking} and {\color{googleyellow}Verification}, respectively from "example quote" and "example quote"

A negative response example, no further explanation is needed at all, SIMPLY return {\color{googlegreen}<NO>}:

{\color{googlegreen}<NO>}
\end{tcolorbox}

\twocolumn[
\clearpage
]

\begin{tcolorbox}[
    halign=flush left,
    breakable,
    colback=googlegray!5!white,
    colframe=googlegray!75!black,
    title={\Large\textbf{MathVista Parser Prompt}},
    fonttitle=\bfseries,
    fontupper=\ttfamily,
    width=\textwidth,     enhanced jigsaw
]
{\color{googlered}Please read the following example. Then extract the answer from the model response and type it at the end of the prompt.}

{Hint: Please answer the question requiring an integer answer and provide the final value, e.g., 1, 2, 3, at the end.
Question: Which number is missing?

Model response: The number missing in the sequence is 14.}

{\color{googlegreen}Extracted answer: 14}

{Hint: Please answer the question requiring a floating-point number with one decimal place and provide the final value, e.g., 1.2, 1.3, 1.4, at the end.
Question: What is the fraction of females facing the camera?

Model response: The fraction of females facing the camera is 0.6, which means that six out of ten females in the group are facing the camera.}

{\color{googlegreen}Extracted answer: 0.6}

{Hint: Please answer the question requiring a floating-point number with two decimal places and provide the final value, e.g., 1.23, 1.34, 1.45, at the end.
Question: How much money does Luca need to buy a sour apple candy and a butterscotch candy? (Unit: \$)

Model response: Luca needs \$1.45 to buy a sour apple candy and a butterscotch candy.}

{\color{googlegreen}Extracted answer: 1.45}

{Hint: Please answer the question requiring a Python list as an answer and provide the final list, e.g., [1, 2, 3], [1.2, 1.3, 1.4], at the end.
Question: Between which two years does the line  graph saw its maximum peak?

Model response: The line graph saw its maximum peak between 2007 and 2008.}

{\color{googlegreen}Extracted answer: [2007, 2008]}

{Hint: Please answer the question and provide the correct option letter, e.g., A, B, C, D, at the end.
Question: What fraction of the shape is blue?Choices: (A) 3/11 (B) 8/11 (C) 6/11 (D) 3/5

Model response: The correct answer is (B) 8/11.}

{\color{googlegreen}Extracted answer: B}

{\color{googleblue}<query><response>}
\end{tcolorbox}

\end{document}